\documentclass[conference]{IEEEtran}
\IEEEoverridecommandlockouts

\usepackage{cite}
\usepackage{amsmath,amssymb,amsfonts}
\usepackage{algorithmic}
\usepackage{graphicx}
\usepackage{textcomp}
\usepackage{xcolor}
\usepackage{booktabs}
\usepackage{multirow}
\usepackage{url}
\usepackage{xspace}
\usepackage{caption}

\usepackage[hidelinks]{hyperref}

\usepackage{fancyhdr}
\fancypagestyle{preprint}{
    \fancyhf{}
    
    \fancyhead[C]{\footnotesize \textsf{Preprint. Accepted for publication in IJCNN 2026. © 2026 IEEE.}}
}
\newcommand{\method}{Predict-then-Diffuse\xspace}
\newcommand{\short}{AdaRLP\xspace}
\newcommand{\token}[1]{\texttt{<#1>}\xspace}
\newcommand{\dllm}{D-LLM\xspace}

\def\BibTeX{{\rm B\kern-.05em{\sc i\kern-.025em b}\kern-.08em
    T\kern-.1667em\lower.7ex\hbox{E}\kern-.125emX}}

\begin{document}

\title{Predict-then-Diffuse: Adaptive Response Length for Compute-Budgeted Inference in Diffusion LLMs}

\author{
\IEEEauthorblockN{Michael Rottoli, Subhankar Roy, Stefano Paraboschi}
\IEEEauthorblockA{
\textit{Università degli Studi di Bergamo, Italy}\\
}
\thanks{© 2026 IEEE. Preprint. Accepted for publication in IJCNN 2026. Personal use of this material is permitted. Permission from IEEE must be obtained for all other uses, in any current or future media, including reprinting/republishing this material for advertising or promotional purposes, creating new collective works, for resale or redistribution to servers or lists, or reuse of any copyrighted component of this work in other works. }
}

\maketitle

\thispagestyle{preprint}

\begin{abstract}
Diffusion-based Large Language Models (D-LLMs) represent a promising frontier in generative AI, offering fully parallel token generation that can lead to significant throughput advantages and superior GPU utilization over the traditional autoregressive paradigm. However, this parallelism is constrained by the requirement of a fixed-size response length prior to generation. This architectural limitation imposes a severe trade-off: oversized response length results in computational waste on semantically meaningless padding tokens, while undersized response length causes output truncation requiring costly re-computations that introduce unpredictable latency spikes. To tackle this issue, we propose \method, a simple and model-agnostic framework that enables compute-budgeted inference per input query by first estimating the response length and then using it to run inference with D-LLM. At its core lies an Adaptive Response Length Predictor (\short), which estimates the optimal response length given an input query. As a measure against under-estimating the response length and re-running inference with a higher value, we introduce a data-driven safety mechanism based on a small increase of the predicted length. As a whole, our framework avoids wasting computation on padding tokens, at the same time preserving output quality. Experimental validation on multiple datasets demonstrates that \method significantly reduces computational costs (FLOP) compared to the default \dllm inference mechanism, while being robust to skewed data distributions.
\end{abstract}


\section{Introduction}
The landscape of Natural Language Processing (NLP) is dominated by highly-scalable Autoregressive Large Language Models (AR-LLMs) such as GPT \cite{brown2020language} and LLaMA \cite{touvron2023llama}. These models generate text sequentially, a process that inherently limits inference speed and hardware utilization due to memory-bandwidth bottlenecks and especially due to its sequential nature. Recently, Diffusion-based Language Models (D-LLMs) like LLaDA \cite{nie2025large} and Dream \cite{ye2025dream7b} have emerged as a promising alternative. By adapting diffusion principles from image generation \cite{ho2020denoising}, D-LLMs generate all tokens in an output simultaneously using bidirectional attention, thus departing from sequential generation. A crucial advantage of the diffusion paradigm is that it unlocks the potential for high-throughput generation with superior GPU utilization.

\begin{figure*}
    \centering
    \includegraphics[width=0.8\linewidth]{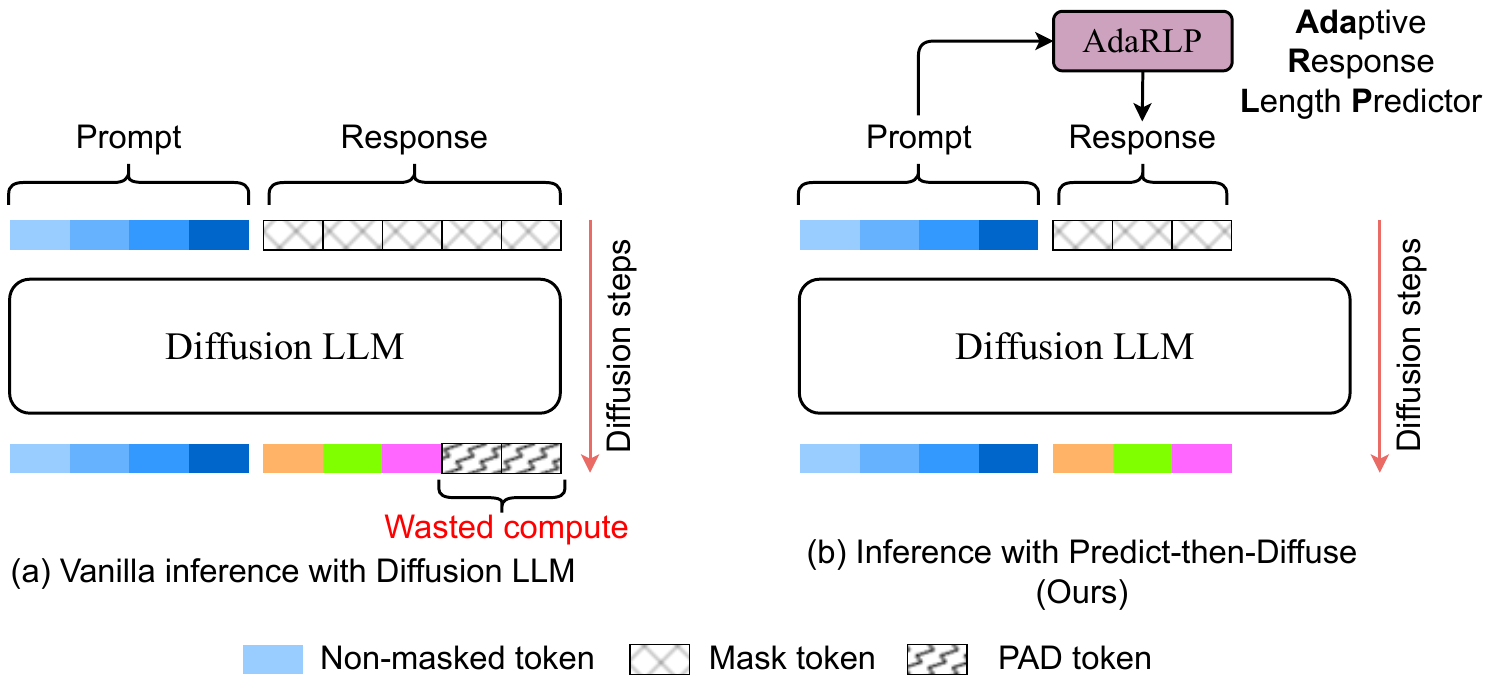}
    \caption{\textbf{Comparison of inference strategies with Diffusion LLMs.} (a) Vanilla inference with a fixed response length results in wasted compute for the \texttt{<PAD>} tokens. (b) In our proposed \method system, the \textbf{Ada}ptive \textbf{R}esponse \textbf{L}ength \textbf{P}redictor (\textbf{\short}) auxiliary module predicts the response length conditioned on the input prompt, circumventing wasted compute on processing \texttt{<PAD>} tokens, without affecting the quality of the response}
    \label{fig:pipeline}
\end{figure*}

Despite the parallelization promises offered by D-LLMs, they face an operational hurdle of \textit{static sequence length} during inference. Unlike AR models that for a given input query generate output until an \token{EOS} token is produced, a \dllm must be initialized with a fixed sequence (or response) length, defining the length of the output sequence. As illustrated in Fig. \ref{fig:pipeline} (a), a \dllm performs computations over both meaningful tokens and semantically meaningless \token{PAD} tokens because the model cannot know \textit{a priori} at the start of the diffusion process which tokens will be meaningful. The waste becomes substantial as the true response length and maximum response length diverge, since the computational complexity in Transformer-based models scales quadratically with the sequence length. Moreover, manually setting the response length to a conservatively low value causes truncation of the output, degrading quality and user experience. 

Since queries vary significantly in their complexity, existing mitigations, such as Block Diffusion \cite{arriola2025block}, introduce a semi-autoregressive approach, which combines the AR and diffusion paradigms to generate arbitrary-length outputs. While this enables intra-block parallel sampling, it calls for modified training objective and sampling algorithm. Distinguishing ourselves from previous approaches, we propose the \textbf{\method} framework, which fulfills the same goal without making any architectural modifications or introducing any new training objectives. Our proposed framework is based on the simple idea of predicting the optimal response length conditioned on the input prompt, prior to generation, such that computation on \token{PAD} tokens can be minimized. At its core is an \textbf{Ada}ptive \textbf{R}esponse \textbf{L}ength \textbf{P}redictor (\textbf{\short}) module (see Fig. \ref{fig:pipeline} (b)) that implements a lightweight predictive model, which takes as input the raw input prompt and predicts the response length. Its lightweight design adds negligible computational overhead to the actual output generation process and its model-agnostic nature allows it to be plugged into any \dllm. In addition, we equip \method with a data-driven risk-aware Safety Margin mechanism that mitigates the issue of under-predicting the response length. This reduces the degradation of the output quality and provides a reliable estimate of latency, which is crucial in production environments.

In summary, our \textbf{contributions} are:

\begin{itemize}
    \item We highlight and study the computational bottleneck of standard Diffusion LLMs during inference. To this end we derive an analytical cost model for \dllm inference and validate it empirically on modern hardware.
    \item We propose \textbf{\method} framework that estimates the response length prior to generation, thus reducing the computational waste on \token{PAD} tokens. In addition, we introduce a data-driven Safety Margin that trades a negligible padding overhead for preventing output truncation.
    \item We benchmark our proposed framework and a series of baselines on a collection of multiple datasets. Experimental results demonstrate that \method reduces FLOP by 99.34\% with respect to default \dllm inference, while maintaining quality of output.
\end{itemize}

\section{Related Work}

\noindent \textbf{Autoregressive (AR) LLMs} are a family of generative models that model the conditional probability of each token given all the previous tokens, as exemplified by GPT and LLaMA  \cite{brown2020language,touvron2023llama}. Albeit highly-scalable and effective, AR generation is inherently sequential (i.e., predict one token at a time), creating a bottleneck during inference. To improve throughput and efficiency in AR, inference engines (e.g., vLLM \cite{kwon2023efficient}) have been developed that primarily improve KV-cache memory management. In order to make LLMs inherently parallel, an alternative diffusion-based language model has been studied, which is the focus of this work.

\vspace{1mm}
\noindent\textbf{Diffusion LLMs (D-LLMs)} \cite{li2025survey} are built upon the idea of diffusion, where clean text is systematically corrupted (through noise injection or masking) and then the model learns to remove the noise through iterative denoising (e.g., LLaDA \cite{nie2025large}). By design, D-LLMs are parallelizable on modern hardware at inference, as the bidirectional attention allows all token positions to be simultaneously updated at each denoising step. However, D-LLMs suffer from two main drawbacks: (i) D-LLMs are less performant than AR models, and (ii) Unlike AR LLMs, D-LLMs must be initialized with a fixed response length \textit{a priori} during inference. In particular, fixed response length can be problematic, as it can lead to computation being wasted on processing \texttt{<PAD>} tokens, and output being truncated if the true response is longer than the maximum response length. This has led to further studies aiming to address these two challenges.

\vspace{1mm}
\noindent\textbf{Budgeted-inference of LLMs}. To combine the strengths of both approaches, BlockDiffusion \cite{arriola2025block,sahoo2024simple} unifies the two paradigms by modelling an autoregressive probability distribution over blocks of discrete random variables. This allows the model to generate arbitrary-length high-quality output and also enables intra-block parallel sampling. In contrast, our proposed compute-budgeted strategy makes no modifications to the architecture or the training objective of a D-LLM. We train a lightweight \short module that predicts the expected response length for a given input query, thus saving computation on processing \texttt{<PAD>} tokens. A related work, TimeBill \cite{fan2025timebilltimebudgetedinferencelarge}, aims to achieve an optimal KV cache eviction ratio through the estimation of end-to-end execution time, but unlike our work, TimeBill employs it for AR LLMs that are time-budgeted during inference.

\section{Methods}
\label{sec:methods}
In this section, we present \method, a framework designed for compute-budgeted inference. We first discuss some preliminaries on D-LLMs (Sec. \ref{sec:prelim}), and the analytical cost model of D-LLMs (Sec. \ref{sec:cost}). Finally, we present the \method framework (Sec. \ref{sec:framework}) in detail.

\subsection{Preliminaries: Diffusion LLMs}
\label{sec:prelim}
Diffusion models, pioneered for image generation tasks \cite{ho2020denoising}, have also been used for language modelling tasks. At its core, D-LLMs learn to recover data, i.e., text, from increasingly noised versions by reverse diffusion process, and generate new samples by iteratively reversing the stochastic corruption. Based on the noising process, which can be performed either in the continuous embedding space or in the discrete token space, D-LLMs can be categorized as Continuous D-LLMs and Discrete D-LLMs, respectively. Our proposed framework can be seamlessly applied to any \dllm, but for convenience we focus on the Discrete D-LLM LLaDA \cite{nie2025large}.

In LLaDA, the \textbf{forward diffusion} process masks a clean sequence $x_0$ with some probability $t \sim \mathcal{U}(0, 1)$ with \texttt{<MASK>} tokens to obtain a partially masked sequence $x_t$. The \textbf{reverse diffusion} process learns to reverse masking by predicting the tokens in masked positions. In detail, a parametric model $p_\theta(\cdot \mid x_t)$ takes the masked sequence $x_t$ as input, and predicts all the masked tokens, denoted by $\mathbf{M}$, simultaneously. The model is trained with a cross-entropy loss:
\begin{equation}
    \label{eqn:llada-loss}
    \mathcal{L}(\theta) = - \mathbb{E}_{t, x_0, x_t} \Big[\frac{1}{t} \sum^{L}_{i=1} \mathbf{1} [x^i_t = \mathbf{M}]\log p_\theta(x^i_0 \mid x_t)] \Big],
\end{equation}
where $\mathbf{1[\cdot]}$ is an indicator function that computes loss only over the masked tokens, and $L$ is the sequence length. During inference, starting from a completely masked sequence ($t=1$), the model predicts a complete sequence of tokens. Inference is iterative, as it involves re-masking some tokens and then gradually refining its prediction as $t$ moves from 1 to 0. Notably, because of fixed response length $L$, the tokens appearing after the \texttt{<EOS>} token, designated as \texttt{<PAD>} tokens, are discarded after the final output is generated.

\subsection{Analytical Cost Model}
\label{sec:cost}

In this work, we highlight that in LLaDA the \token{PAD} tokens, which have no impact on the final generated output, waste computation during inference, a phenomenon not present in AR LLMs. To better understand the wasted computation in \dllm, we first quantify the computation during inference by deriving its analytical cost model. We use a standard FLOP (FLoating point OPeration) accounting for Transformer models \cite{hsu2024jax, kaplan2020scalinglawsneurallanguage} (see Tab. \ref{tab:flops} for per-component computation) to calculate the FLOP of a forward pass for a block $l$:

\begin{equation}
    \text{FLOP}_{l} = \underbrace{(6DF + 8D^2)L}_{\text{Terms linear in } L} + \underbrace{(4D)L^2}_{\text{Term quadratic in } L}.
\end{equation}

For a \dllm with $N$ Transformer blocks and $T$ diffusion steps, the total FLOP during inference are:

\begin{equation}
    \label{eq:cost_model}
    \text{FLOP}_\text{total} = T \cdot N \cdot D \cdot (\alpha L + \beta L^2),
\end{equation}
where we use $\alpha = 6F + 8D$ and $\beta = 4$ for brevity. For LLaDA-8B ($D=4096, F\approx 3D$), the quadratic term dominates when $L > \alpha/\beta \approx 26,000$. Ideally, for short sequences the cost grows linearly, whereas for longer sequences, the quadratic attention term dominates the cost. In the case of \dllm, the sequence length $L$ needs to be fixed upfront, with the aim of catering to variable-length responses. As a consequence, when the actual response length is short, significant computation is wasted on processing the \token{PAD} tokens that constitute most of the response. 

\subsection{Predict-then-Diffuse Framework}
\label{sec:framework}

We propose \textbf{\method}, an intuitively simple framework designed for compute-budgeted inference with a \dllm. The main idea of \method lies in predicting the expected response length given an input prompt. In most cases, this produces a significant shortening of the sequence length, and therefore a reduction in computation on processing the \token{PAD} tokens. It operates in the pre-inference stage and incurs negligible computational overhead with respect to the FLOP incurred by the actual inference with \dllm.

\begin{table}[t]
    \caption{Per-block FLOP for major components of \textbf{Gated Transformer with Multi Head Attention} during a single \textbf{forward pass} (inference-only) as a function of sequence length ($L$), hidden dimension ($D$), and MLP width ($F$)}
    \label{tab:flops}
    \begin{center}
    \begin{tabular}{l|c}
        \toprule
        \textbf{Component} & \textbf{Computational Cost (FLOP)} \\
        \midrule
        MLP block & $6LDF$ \\
        Q, K, V, O projections & $8LD^2$ \\
        Dot-product attention & $4DL^2$ \\
        \bottomrule
    \end{tabular}
    \end{center}
\end{table}

As shown in Fig. \ref{fig:pipeline}(b), at the core of \method is the \textbf{Ada}ptive \textbf{R}esponse \textbf{L}ength \textbf{P}redictor (\textbf{\short}) module that takes as input a prompt and outputs the expected response length. We cast the response length prediction as a regression task. Formally, we assume access to a paired dataset $\mathcal{D} = \{(s_i, k_i)\}^n_{i=1}$, where $s_i \in \mathcal{S}$ is raw text or the $i-$th input prompt in the dataset, and $k_i \in \mathcal{Z}$ is the response length of its corresponding output. Using $\mathcal{D}$, we train the \short module, or a function $f_\text{\short}: \mathcal{S} \mapsto \mathcal{Z}$, to predict the response length given a prompt. Note that $\mathcal{D}$ is not any specialized dataset, but any supervised fine-tuning (SFT) dataset that is typically used for instruction after the pre-training phase.

\vspace{1mm}
\noindent \textbf{\short design}. We design \short with the following considerations. \textit{First}, we do not want to introduce any significant computational overhead in predicting response length. \textit{Second}, we want to keep the design fairly simple, with no need for an extensive ad-hoc feature engineering. With these goals in mind, we employ, as the predictor $f_\text{\short}$, CatBoost \cite{prokhorenkova2018catboost}, a widely used machine learning algorithm based on gradient boosting over decision trees. CatBoost offers two distinct advantages that satisfy our design desiderata: (i) its extremely fast inference speed ($< 1$ms), (ii) its ability to natively handle mixed data types, including raw text, in a robust way, and (iii) its consistently high performance on tabular and heterogeneous data without extensive tuning; all these features make it an extremely good response-length predictor.

To test this hypothesis, we evaluated two approaches: (i) $\textbf{\short}_\text{text-only}$: the raw prompt string is passed directly to CatBoost, utilizing its internal quantization and support for text features. (ii) $\textbf{\short}_\text{engineered}$: a more engineered model that in addition to the raw prompt string, uses metadata such as token counts, specific keywords (e.g., "summarize", "list"), punctuation patterns, and Shannon entropy. As we will show later in Sec. \ref{sec:predictive_performance}, $\textbf{\short}_\text{text-only}$ significantly outperformed $\textbf{\short}_\text{engineered}$, suggesting that the native CatBoost algorithm is sufficient to handle a simple task of predicting response length. In theory, one can adapt a small language model to predict response length, but at the cost of increased inference time; hence, we did not explore this option.

\vspace{1mm}
\noindent \textbf{Introducing Safety Margin}. Contrary to over-predicting the response length, one potential drawback of under-predicting the expected response length is that it will lead to output truncation, i.e., incomplete response from \dllm, which in turn impacts the output quality. A straightforward solution to this problem consists in increasing the response length upon failure, and re-invoking the \dllm for inference, but this approach introduces additional FLOP and wastes computation. 

As a more adaptive safeguard, we introduce a data-driven safety mechanism specifically designed to prevent degradation of the quality of output. Let us denote the predicted response length from $f_\text{\short}$ by $\hat{L}$. We introduce a safety margin $\delta$ that is added to the initial prediction $\hat{L}$. The margin is determined empirically, set by calculating a high percentile (e.g., the 95th, denoted $p_\text{safe}$) of the positive prediction errors, i.e., instances where true length is greater than the predicted length, $\hat{L} < k$:

\begin{equation}
\delta = \text{Quantile}_{p_\text{safe}}[\{ k_i - \hat{L}_i \mid k_i > \hat{L}_i \}].
\end{equation}

The final effective response length $L^*$ passed to the D-LLM:
\begin{equation}
    L^* = \min(\hat{L} + \delta, L_\text{max}),
\end{equation}
where $L_\text{max}$ is the maximum response length \dllm was trained with.
This statistical buffer minimizes fallbacks while maintaining a tight and close-to-optimal response length.

\vspace{1mm}
\noindent\textbf{Inference with \method}. Our proposed framework is agnostic to the implementation of the \dllm, as it operates in the pre-inference phase. While in this work we experiment only with LLaDA, the proposed framework can be seamlessly integrated into the inference pipeline of any \dllm. In detail, once $L^*$ is determined for a given test sample, LLaDA inference is invoked with the response length $L^*$, without needing to alter its native inference mechanism. LLaDA then proceeds with its standard $T$ denoising steps, but only processes a sequence of effective length $L^*$. The experiments show that this drastically reduces the computational load for shorter output responses. In very extreme cases of the model failing to generate a \token{EOS} token even with utilizing $L^*$ as the effective length, we adopt a conservative fallback strategy of re-running inference with maximum response length $L_{\text{max}}$.

\section{Experiments}

\subsection{Empirical Validation of Analytical Cost Model}
\label{sec:validation}

We first empirically validate the analytical cost model introduced in Sec. \ref{sec:cost}, where we theoretically showed that FLOP scale quadratically with the response length $L$ of a \dllm, especially for long sequence lengths. To validate this, we run the LLaDA-8B model with varying response length on NVIDIA H100 (80 GB VRAM) GPUs. In detail, we measured the total FLOP using DeepSpeed \cite{aminabadi2022deepspeedinferenceenablingefficient} and the VRAM usage for a single generation (diffusion steps $T=128$) with response length ranging from 64 to 30k tokens.

\begin{figure}[t]
    \centerline{\includegraphics[width=0.95\columnwidth]{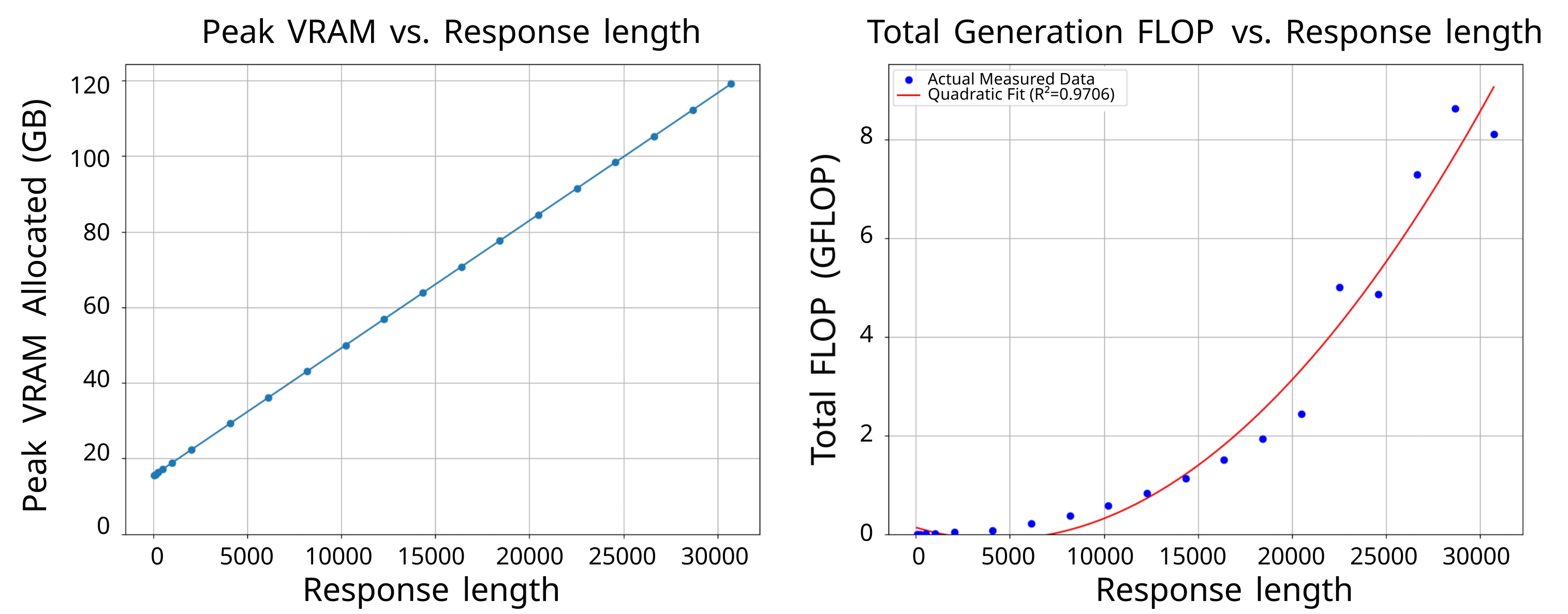}}
    \caption{Empirical validation of D-LLM cost scaling on NVIDIA H100 GPUs. \textbf{Left:} VRAM usage scales linearly. \textbf{Right:} Total FLOP scales quadratically with canvas size ($R^2=0.9706$), strongly supporting the theoretical structure ($O(\alpha L + \beta L^2)$) of our analytical cost model.}
\label{fig:validation}
\end{figure}

\vspace{1mm}
\noindent\textbf{Results}. We observe from Fig. \ref{fig:validation} (\textbf{left}) that VRAM usage is linear with the response length. In Fig. \ref{fig:validation} (\textbf{right}) we observe that FLOP follow a linear fit when the response length is short, whereas for a longer response length, FLOP exhibit a quadratic fit. Overall, the curve follows a quadratic non-linear trend (variance explained, $R^2 = 0.9706$). This experiment confirms that the growth term is not just a theoretical construct, but a dominant factor in real-world deployments where the response length can vary widely.

\subsection{Experimental Setup}
\label{sec:setup}

\noindent\textbf{Benchmarks}. We aggregated 39,994 prompt-response pairs from 5 diverse public sources: ShareGPT, Alpaca, Dolly-15k, OpenOrca, and ELI5 to benchmark our framework \cite{xia-etal-2025-ground, chen2024alpagasus, zhang2025instructiontuninglargelanguage, wang2024openchat, mitra2023orca2teachingsmall}. 
We curated this composite benchmark to ensure a high variance in output response lengths ($mean \approx 96$, $ std \approx 120,$ $kurtosis \approx 107$), with an emphasis on long-form generation. This selection strategy was critical to mitigate the bias toward short responses typical of standard open-source instruction-tuning datasets, thereby providing a robust testbed for adaptive response length.
Responses were tokenized using the GPT2TokenizerFast tokenizer before calculating the length. The dataset was split 80/20 for training and testing. 

\vspace{1mm}
\noindent\textbf{Models and baselines}. As discussed in Sec. \ref{sec:framework}, we evaluated two variants of CatBoost models for implementing \short: (i) $\textbf{\short}_\text{text-only}$ that uses CatBoost's native support for text features, and (ii) $\textbf{\short}_\text{engineered}$ that additionally uses hand-engineered features. The predicted real-valued output length in tokens is rounded to the closest integer, since response length is an integer value.

Due to the lack of baselines and competitor methods, we construct the following baselines for quantitative comparison: (i) \textbf{Max Response Length}, where we set $L=L_{\text{max}}=4096$, default setting used by LLaDA. (ii) \textbf{Static Response Length} ($L=200$), a static-length conservative baseline that doubles on fallback if output truncation occurs. (iii) \textbf{Mean Doubling Heuristic}, a strategy that initializes the response length to the mean dataset length ($\approx 95$ tokens) and doubles the size ($95 \to 190 \to \dots$) if truncation occurs. (iv) \textbf{Oracle}, where we assume knowledge of the true response length, i.e., $L = k$. Unless otherwise stated, all proposed models and baselines use LLaDA-8B for output generation.

\vspace{1mm}
\noindent\textbf{Metrics}. We compare different methods mainly using Total Tera Floating-point Operations (\textbf{TFLOP}) on the full test set, including penalties for fallback re-runs.

\begin{table}[t]
    \caption{Length-prediction metrics on the held-out test set. $\textbf{\short}_\text{text-only}$ outperforms the $\textbf{\short}_\text{engineered}$ variant with hand-engineered features}
    \label{tab:metrics}
    \begin{center}
    \begin{tabular}{l|ccc}
        \toprule
        \textbf{Model}&\multicolumn{3}{c}{\textbf{Metrics}} \\
        & \textbf{\textit{RMSE}}& \textbf{\textit{MAE}}& \textbf{\textit{\% $\le$ 10\% error}} \\
        \midrule
        $\textbf{\short}_\text{engineered}$ & 81.6 & 51.6 & 10.5 \\
        $\textbf{\short}_\text{text-only}$ & \textbf{11.4} & \textbf{1.7} & \textbf{97.5} \\
        \bottomrule
    \end{tabular}
    \end{center}
\end{table}

\subsection{Predictor Performance}
\label{sec:predictive_performance}
\begin{figure}[t]
    \centerline{\includegraphics[width=0.95\columnwidth]
    {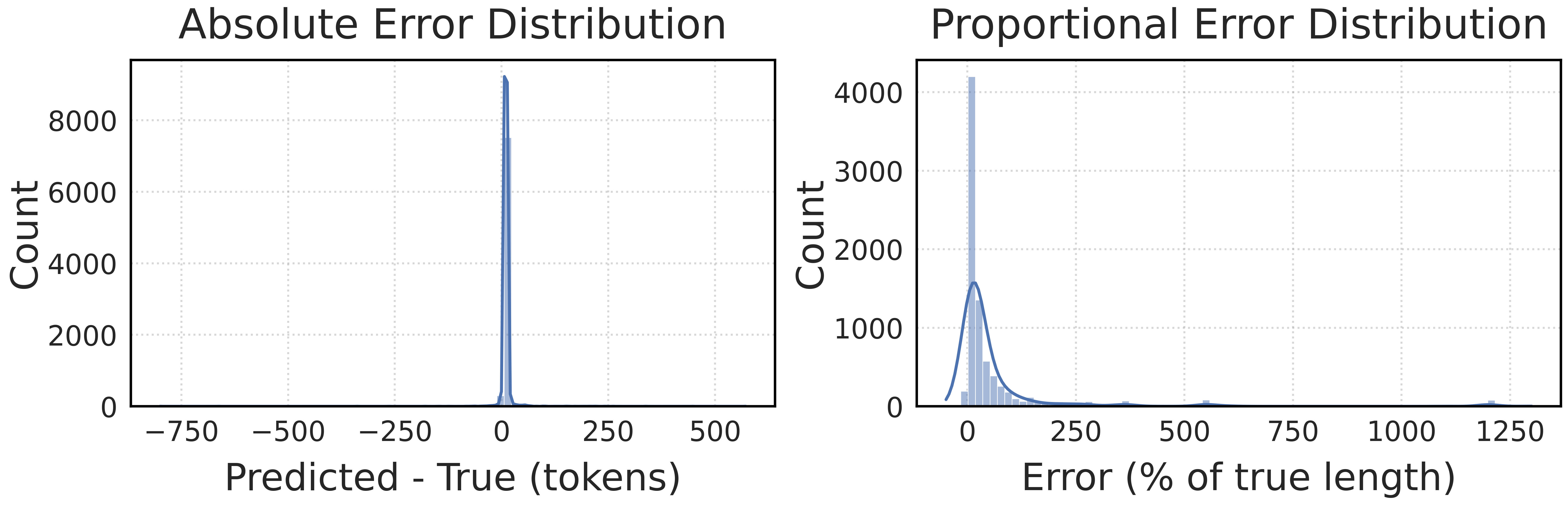}}
    \caption{Absolute and Proportional error distributions of $\textbf{\short}_\text{text-only}$}
    \label{fig:error_plots}
\end{figure}

\begin{figure}[t]
    \centerline{\includegraphics[width=0.95\columnwidth]
    {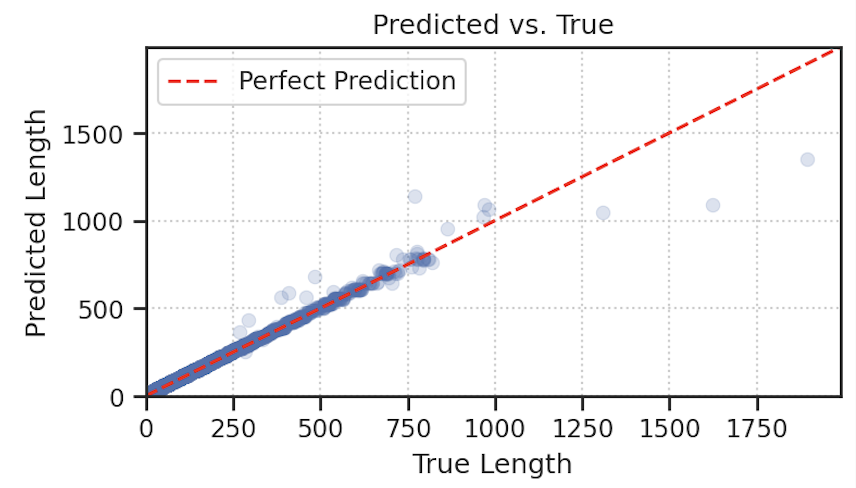}}
    \caption{Deviations from the perfect prediction of $\textbf{\short}_\text{text-only}$}
    \label{fig:length_accuracy}
\end{figure}

We compare the response length predictive performance of two variants of \short by comparing their Mean Absolute Error (MAE) and Root Mean Square Error (RMSE). As evident from Tab. \ref{tab:metrics}, the simpler $\textbf{\short}_\text{text-only}$ outperformed the feature-engineered $\textbf{\short}_\text{engineered}$ model by a large margin. In detail, $\textbf{\short}_\text{text-only}$ achieved a MAE of 1.7 tokens and an RMSE of 11.4 tokens, with 97.5\% of predictions falling within a 10\% error margin (see Fig. \ref{fig:error_plots}). This indicates that CatBoost effectively learns length-correlated features (e.g., ``write a poem" vs. ``write a novel") directly from raw text (see Fig. \ref{fig:length_accuracy}). Given its clear superiority, the remainder of our analysis focuses on $\textbf{\short}_\text{text-only}$.

\subsection{Computational Efficiency}
Tab. \ref{tab:savings} summarizes the computational costs of \textbf{\method} and the baselines introduced in Sec. \ref{sec:setup}. The \textbf{Max Response Length} baseline is prohibitively expensive, as evident from the highest computational cost of 4.3 TFLOP, due to a long response length, with almost every alternative showing significant improvements. This again underscores the need for adaptive response length during \dllm inference.

\begin{table}[t]
    \caption{Comparison of Computational Cost (in TFLOP), Savings and Fallback rate}
    \begin{center}
    \resizebox{\linewidth}{!}{
    \begin{tabular}{l|ccc}
        \toprule
        \textbf{Methods} & \textbf{TFLOP} & \textbf{Savings (\%)} & \textbf{Fallback Rate (\%)} \\
        \midrule
        Max Response Length (4096) & 4.03 & 0.0\% & 0.0\% \\
        Static Response Length (200) & 0.054 & 98.6\% & 22.1\% \\
        Doubling Heuristic & 0.027 & 99.31\% & - \\
        \textbf{Predict-then-Diffuse} (Ours) & \textbf{0.026} & \textbf{99.34\%} & \textbf{0.1\%} \\
        \midrule
        \textcolor{gray}{Oracle} & \textcolor{gray}{0.024} & \textcolor{gray}{99.4\%} & \textcolor{gray}{0.0\%} \\
        \bottomrule
    \end{tabular}
    }
    \label{tab:savings}
    \end{center}
\end{table}

Interestingly, the \textbf{Mean Doubling Heuristic} baseline is competitive (99.31\% savings) with \textbf{\method}. This is due to the specific distribution of our dataset (mean length $\approx$ 95, median $\approx$ 63). Since most answers are short, starting at the mean covers the majority of cases without fallback. However, \textbf{Predict-then-Diffuse} still slightly outperforms the heuristic-based baseline (99.34\% savings) while offering superior robustness. The heuristic is brittle, as on a bimodal dataset (e.g., 50\% short chats, 50\% long reports) the mean-based initialization would fail frequently for long prompts, incurring higher costs. 
To verify this, we simulated a dataset with responses following a bimodal distribution (60\% short queries, $mean=50$ tokens; 40\% long reports, $mean=3000$ tokens). We observe that the heuristic-based baseline frequently fails multiple times before converging on the required length (e.g., $95 \to 190 \to \dots \to 4096$), incurring cumulative retry costs. In contrast, \textbf{\method} retains a 19\% computational advantage over the heuristic in this bimodal regime, demonstrating its necessity for general-purpose deployment where output length variance is high.

\subsection{Latency Determinism}
Beyond aggregate TFLOP reduction, reliable inference entails another important requirement: \textit{latency determinism}.
While heuristic strategies may achieve high average savings on skewed distributions, they also introduce stochastic latency behavior and exhibit a high variance. A long-tail query (e.g., a lengthy report) processed by the \textbf{Mean Doubling Heuristic} baseline triggers multiple inferences (e.g., $L=95 \to 190 \to \dots$), causing the user-perceived latency to frequently spike unpredictably. 
In contrast, \textbf{\method} aims to yield a consistent, single-shot generation profile for 99.9\% of requests. By integrating the data-driven Safety Margin ($\delta=5$ tokens), as discussed in Sec. \ref{sec:framework}, we suppressed the frequency of fallback retries to negligible levels (down from $1.43\%$ to $0.1\%$). This ensures that, while the theoretical worst-case latency exists, it is statistically contained, yielding a stable latency profile. Furthermore, the overhead of the \short ($<0.04$ms) is negligible compared to the generation time. Overall, a stable end-to-end execution time is often more critical than raw throughput for meeting strict Service Level Agreements in production environments.

\begin{table}[t]
    \caption{LLaDA Generation Quality across different response lengths. Results taken from \cite{nie2025large}}
    \begin{center}
    \begin{tabular}{l|cc}
    \toprule
    \textbf{Canvas Size} & \textbf{GSM8K Accuracy} & \textbf{HumanEval Pass@1} \\
    \midrule
    1024 & 70.3\% & 35.4\% \\
    512 & 70.8\% & 32.9\% \\
    256 & 70.0\% & 32.9\% \\
    \bottomrule
    \end{tabular}
    \label{tab:quality}
    \end{center}
\end{table}

\subsection{Impact on Generation Quality}
We rely on prior results on whether adaptive sizing affects generation quality. We reference findings from LLaDA \cite{nie2025large}, where the model was evaluated with different response lengths. As reported in Tab. \ref{tab:quality}, performance remains stable across response lengths. This supports our premise that as long as the canvas $L^*$ is sufficient to contain the answer ($L^* \ge k$), the \dllm performs optimally. The degradation comes only from significant truncation, which our system actively prevents via the Safety Margin and fallback mechanisms.

We observe that D-LLMs exhibit the ability to adapt verbosity based on the available response length. As illustrated in Fig. \ref{fig:llm_adaptation_table}, when constrained to a short response length, the model tends to produce more concise yet correct answers, whereas a longer response length results in more detailed reasoning. 

\begin{figure}[t]
    \centering
    \fbox{
        \begin{minipage}{0.9\columnwidth}
        \centering
        \textbf{Input Prompt:} \\
        \texttt{"Lily runs 12 km/h for 4 hours, then 6 km/h. How far does she run in 8 hours?"}
        \end{minipage}
    }
    \vspace{0.2cm}
    \begin{center}
    \begin{tabular}{p{0.45\columnwidth} p{0.45\columnwidth}}
        \toprule
        $L=24$ tokens & $L=64$ tokens \\
        \midrule
        \textbf{LLaDA Output:} \textit{Lily can run 48 + 24 = 72 kilometers in 8 hours.}
        &
        \textbf{LLaDA Output:} \textit{In the first 4 hours, Lily runs 12 x 4 = 48 kilometers. Then, she runs 6 x 4 = 24 kilometers. In total, Lily runs 48 + 24 = 72 kilometers.} \\
        \bottomrule
    \end{tabular}
    \end{center}
    \caption{Adaptability to response length ($L$) constraint. For the same prompt, LLaDA generates outputs of varying verbosity and token count while maintaining answer correctness}\label{fig:llm_adaptation_table}
\end{figure}

\section{Conclusions and Future Work}

In this work, we studied the computational costs associated with Diffusion LLMs (e.g., LLaDA) that operate with a fixed response length during inference. Due to its implicit design, when the true response length is shorter than the fixed response length, compute is wasted processing semantically meaningless \token{PAD} tokens. As a remedy, we introduced \method, a framework designed for compute-budgeted inference with \dllm. The core idea lies in predicting the response length conditioned on the input prompt using a lightweight and fast \short module. We also introduced a data-driven Safety Margin that circumvents output truncation, and therefore minimizing frequent fallbacks and ensuring predictable inference latency. Experiments on a collection of several datasets show that the proposed framework brings significant savings in computation when compared with default \dllm inference mechanisms and other baselines based on heuristics. As D-LLMs with massive context window (e.g., 32k or 128k tokens) become more prevalent, the benefits of the \method will become more significant.

A limitation of departing from fixed-size response length is that it prevents batching during inference. Although a straightforward solution to this problem involves grouping samples with similar estimated response lengths, more sophisticated and optimized solutions \cite{kwon2023efficient} are left for future work.

\section*{Code Availability}
Our source code and experimental setup are available at: \url{https://github.com/mchl-labs/predict-then-diffuse}

\section*{Acknowledgments}
The authors acknowledge the ISCRA initiative of CINECA and project EVFTEG funded by MASE “Mission Innovation 2.0 - D.M. 386/2023”.

\bibliographystyle{IEEEtran}
\bibliography{references} 

\end{document}